\documentclass[11pt]{article}
\usepackage[T1]{fontenc}
\usepackage[utf8]{inputenc}
\usepackage{lmodern}
\usepackage[margin=1in]{geometry}
\usepackage{amsmath,amssymb,booktabs,array,longtable,graphicx,microtype}
\usepackage{xcolor,tikz,tikz-cd,titlesec}
\usetikzlibrary{arrows.meta,positioning}
\usepackage[nodisplayskipstretch]{setspace}
\usepackage{etoolbox}
\usepackage[hang,flushmargin]{footmisc}
\usepackage[parfill]{parskip}
\usepackage{caption,fvextra,placeins,needspace}
\usepackage[round,authoryear]{natbib}
\bibpunct{(}{)}{;}{a}{,}{,}

\usepackage[colorlinks=true,linkcolor=blue,citecolor=black,urlcolor=blue,hyperfootnotes=false]{hyperref}
\titleformat*{\section}{\Large\bfseries}
\titleformat*{\subsection}{\large\bfseries}
\AtBeginEnvironment{table}{\singlespacing}
\AtBeginEnvironment{figure}{\singlespacing}
\newcommand{\R}{\mathbb R}
\newcommand{\E}{\mathbb E}
\newcommand{\B}{\mathcal B}
\newcommand{\D}{\mathcal D}
\newcommand{\bb}{\boldsymbol\beta}
\newcommand{\yy}{\mathbf y}

\DefineVerbatimEnvironment{Prompt}{Verbatim}{fontsize=\footnotesize,breaklines=true,frame=single,framesep=3mm}
\graphicspath{{figures/}}
\newcommand{\PaperBlankMAE}{0.6854}
\newcommand{\PaperProfileMAE}{0.4931}

\newcommand{\PaperInitialLookupMAE}{0.3846}
\newcommand{\PaperPtOLookupMAE}{0.3396}

\newcommand{\PaperBOFinalMAE}{0.4252}
\newcommand{\PaperBOFirstLookupMAE}{0.3265}
\newcommand{\PaperBOFinalLookupMAE}{0.2809}
\newcommand{\PaperRandomFinalLookupMAE}{0.3369}
\newcommand{\PaperLinearForwardMAE}{0.2131}
\newcommand{\PaperInitialGPMAE}{0.0766}
\newcommand{\PaperFinalGPMAE}{0.0708}
\newcommand{\PaperFinalBOReduction}{16.6}
\newcommand{\PaperBOFinalProfileReduction}{13.8}
\newcommand{\PaperInitialFitN}{799}
\newcommand{\PaperFinalFitN}{3,299}

\newcommand{\PaperAssignedImprovedN}{329}
\newcommand{\PaperAssignedTiedN}{13}
\newcommand{\PaperAssignedWorsenedN}{158}

\newcommand{\PaperSharedImprovedN}{423}
\newcommand{\PaperSharedTiedN}{11}
\newcommand{\PaperSharedWorsenedN}{66}

\newcommand{\PaperForwardTable}{%
\begin{tabular}{@{}lrrr@{}}
\toprule
Predictor & Training pairs & MAE & RMSE \\
\midrule
Blank / no change & --- & 0.2388 & 0.3220 \\
Linear response & 799 & 0.2131 & 0.2711 \\
Initial GP & 799 & 0.0766 & 0.1055 \\
GP after pass 5 & 3,299 & 0.0708 & 0.0982 \\
\bottomrule
\end{tabular}}
\newcommand{\PaperAssignedTable}{%
\begin{tabular}{@{}lrrrrrr@{}}
\toprule
Method & Valid $n$ & \shortstack{Single run\\/ pass 1} & Pass 2 & Pass 3 & Pass 4 & Pass 5 \\
\midrule
Blank & 500 & 0.6854 & --- & --- & --- & --- \\
Profile prompting & 500 & 0.4931 & --- & --- & --- & --- \\
Linear PtO & 459 & 0.6902 & --- & --- & --- & --- \\
\midrule
\textbf{BO} & 500 & \textbf{0.4202} & \textbf{0.4128} & \textbf{0.4261} & \textbf{0.4256} & \textbf{0.4252} \\
Uniform random & 500 & 0.7296 & 0.7279 & 0.7311 & 0.7271 & 0.7319 \\
\bottomrule
\end{tabular}}
\newcommand{\PaperSharedTable}{%
\begin{tabular}{@{}lrrr@{}}
\toprule
Additional interventions & PtO & BO & Uniform random \\
\midrule
0: initial library & 0.3846 & 0.3846 & 0.3846 \\
500 & 0.3396 & 0.3265 & 0.3627 \\
1,000 & --- & 0.3014 & 0.3527 \\
1,500 & --- & 0.2922 & 0.3488 \\
2,000 & --- & 0.2858 & 0.3442 \\
2,500 & --- & 0.2809 & 0.3369 \\
\bottomrule
\end{tabular}}
\newcommand{\PaperBoundsTable}{%
\begin{tabular}{@{}lrrr@{}}
\toprule
Method / bound & Valid $n$ & MAE (valid) & MAE (matched 459) \\
\midrule
Blank & 500 & 0.6854 & 0.6647 \\
Profile prompting & 500 & 0.4931 & 0.4848 \\
PtO: $\pm0.3$ & 459 & 0.6902 & 0.6902 \\
PtO: $\pm0.2$ & 500 & 0.6252 & 0.6066 \\
PtO: $\pm0.1$ & 500 & 0.5913 & 0.5763 \\
\bottomrule
\end{tabular}}
\newcommand{\PaperRobustnessTable}{%
\begin{tabular}{@{}lrrrrrr@{}}
\toprule
& \multicolumn{3}{c}{Assigned recommendation} & \multicolumn{3}{c}{Shared lookup} \\
\cmidrule(lr){2-4}\cmidrule(l){5-7}
Library & PtO ($n$) & BO & Random & PtO & BO & Random \\
\midrule
1 & 0.6759 (44) & 0.5212 & 0.7512 & 0.3764 & 0.4052 & 0.4344 \\
2 & 0.6931 (49) & 0.4556 & 0.7856 & 0.4020 & 0.3908 & 0.4292 \\
3 & 0.6908 (50) & 0.4940 & 0.7372 & 0.3992 & 0.4032 & 0.4528 \\
4 & 0.7151 (49) & 0.4968 & 0.7624 & 0.3936 & 0.4152 & 0.4428 \\
5 & 0.6609 (45) & 0.5008 & 0.7896 & 0.4156 & 0.4164 & 0.4348 \\
6 & 0.7287 (46) & 0.4976 & 0.7312 & 0.4036 & 0.4064 & 0.4388 \\
7 & 0.6867 (48) & 0.5276 & 0.7768 & 0.4008 & 0.4072 & 0.4400 \\
8 & 0.6687 (46) & 0.4652 & 0.7484 & 0.3848 & 0.3968 & 0.4260 \\
9 & 0.6813 (47) & 0.5000 & 0.7796 & 0.3968 & 0.4148 & 0.4444 \\
10 & 0.7440 (50) & 0.4788 & 0.7444 & 0.3880 & 0.4076 & 0.4388 \\
\midrule
Mean & 0.6945 & 0.4938 & 0.7606 & 0.3961 & 0.4064 & 0.4382 \\
\bottomrule
\end{tabular}}

\title{\LARGE Digital Knockoffs}

\author{Niuniu Zhang\\[0.5em]
\textit{UCLA Anderson School of Management}}
\date{November 20, 2026}

\begin{document}
\maketitle
\begingroup
\renewcommand\thefootnote{}

\footnotetext{Contact:
Niuniu Zhang,
\href{mailto:niuniu.zhang.phd@anderson.ucla.edu}{\texttt{niuniu.zhang.phd@anderson.ucla.edu}}.}
\endgroup
\begin{spacing}{1.05}
\begin{abstract}
\noindent Social simulations often use language model agents to represent people who differ on several measurable traits. However, describing a trait profile in a prompt does not guarantee the agent will express each trait at its intended level. We propose a method that calibrates agents to multidimensional profiles using activation steering: (i) extract steering directions for each trait, (ii) learn a shared mapping from steering strengths to measured trait scores, and (iii) search for a steering intervention for each target. We use Bayesian optimization for this search and evaluate the method on the Big Five personality scores of 500 OpenPsychometrics respondents. In each of five rounds, the assigned interventions yield lower mean absolute error across respondents than the prompting and steering baselines. Further, the model’s responses to a given intervention yield the same trait profile regardless of whose target motivated it, so a poor match for one respondent can be the best match for another. When we match each respondent to the closest profile found across all tested interventions, Bayesian optimization yields 16.6\% lower mean absolute error than uniform random search at the same evaluation budget. Our method thus offers a way to build social agents whose measured traits dovetail with those of the people they seek to represent. Whether this closer fit translates to improved predictions of downstream behavior remains an open question.
\end{abstract}
\end{spacing}
\newpage
\section{Introduction}
Social simulations ask how people with different traits behave and
interact. Language models offer a flexible way to construct the agents in
such simulations: a researcher can describe a person, assign a task, and
observe the model's response \citep{filippas2024homosilicus}.
A difficulty arises before any interaction takes place: how closely does the
resulting agent express the traits it was assigned?

Personality makes this question concrete. A person who completes a Big Five
questionnaire receives five scores for Openness, Conscientiousness,
Extraversion, Agreeableness, and Neuroticism (OCEAN).\footnote{The Big Five
summarize broad dimensions of personality identified through factor analysis
\citep{goldberg1992markers}. The
\href{https://openpsychometrics.org/tests/IPIP-BFFM/}{OpenPsychometrics Big Five test}
offers an online administration of the 50-item IPIP questionnaire.}
An agent can answer the same items and receive the same type of score vector. Suppose the desired profile is
$(4.5,4.9,3.5,3.8,4.6)$ on a 1 to 5 scale, in OCEAN order. A description in a prompt, like ``organized and dependable,''
suggests high conscientiousness but leaves open whether the intended
score is 4.0 or 4.9. Listing the scores in the prompt specifies the
target exactly, but the challenge is getting the agent's own answers to
yield scores close to all five values at once.
\emph{How can we calibrate an LLM agent to such a profile with a
limited budget of model evaluations?}

Activation steering supplies controls for this problem. It shifts internal
representations during inference while leaving the model's weights fixed
\citep{turner2023activation}. Concept directions extracted using existing
methods provide a small set of controls in a high-dimensional model
\citep{beaglehole2026steering}. We extract directions for each OCEAN trait and set their steering strengths with five signed coefficients,
$\bb=(\beta_O,\beta_C,\beta_E,\beta_A,\beta_N)\in\R^5$.

With the model, questionnaire prompt, and extracted directions held fixed, we learn a shared intervention--response map
$\bb\mapsto\yy(\bb)$, where $\yy(\bb)\in\R^5$ gives the measured OCEAN
scores.
Given a person's target profile $\yy_p$, calibration seeks an intervention
$\bb$ whose measured response $\yy(\bb)$ closely matches $\yy_p$.
A single control can affect several trait scores, so the controls must
be chosen jointly.

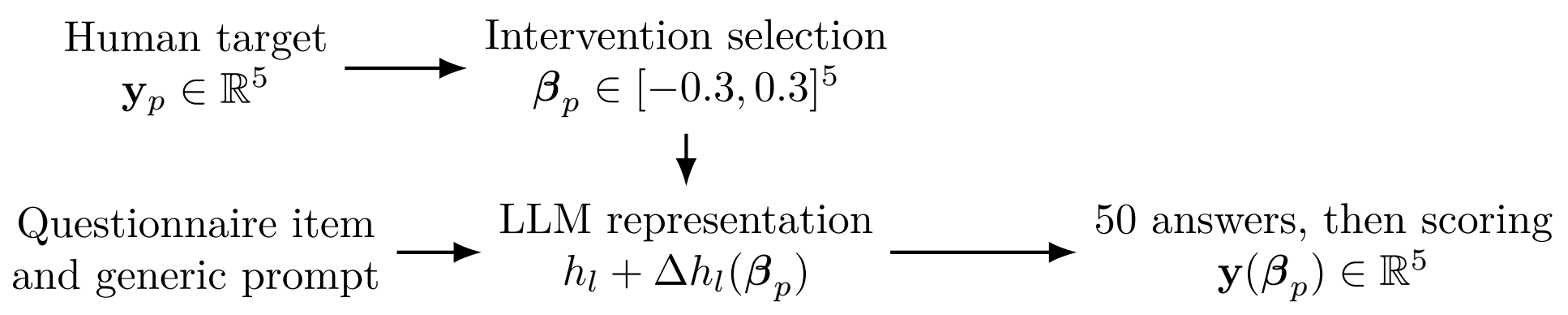
\begin{figure}[htbp]
\centering
\begin{tikzpicture}[>=Latex,every node/.style={align=center,font=\small}]
\node (target) at (0,1.5) {Human target\\$\yy_p\in\R^5$};
\node (select) at (4,1.5) {Intervention selection\\$\bb_p\in[-0.3,0.3]^5$};
\node (prompt) at (0,0) {Questionnaire item\\and generic prompt};
\node (hidden) at (4,0) {LLM representation\\$h_l+\Delta h_l(\bb_p)$};
\node (scores) at (9.2,0) {50 answers, then scoring\\$\yy(\bb_p)\in\R^5$};
\draw[->,thick] (target)--(select);
\draw[->,thick] (select)--(hidden);
\draw[->,thick] (prompt)--(hidden);
\draw[->,thick] (hidden)--(scores);
\end{tikzpicture}
\caption{Joint calibration pipeline.}
\label{fig:calibration}
\end{figure}
Figure~\ref{fig:calibration} illustrates this pipeline. The human target informs
intervention selection. The model then answers each questionnaire item in a
fresh conversation, with its weights and generic prompt unchanged.
Steering changes its activations only; the 50 answers are scored to obtain
$\yy(\bb)$.

As the intervention--response map $\yy$ is fixed, the response to a dose does not depend on whose target motivated it. One observed response can therefore
inform calibration for every person. We use Gaussian processes (GPs) to learn
this shared map and Bayesian optimization (BO) to choose new doses. Every
person retains an individual target and loss; the information used to choose
interventions is shared.

Our contribution fits within a broader effort to build empirical social
agents. \citet{manning2026general} combine theory-grounded instructions with
human data to predict behavior in new games. \citet{jackson2026types} fit
numerical type vectors expressed through prompts to observed human choices,
then evaluate predictions in held-out games. Their type vector is an input
fitted to explain behavior. Here, the human trait vector is an observed
\emph{output target}; the fitted inputs are activation strengths that make
the model's questionnaire scores match it.

Closely related work establishes numerical and multidimensional control.
\citet{fan2026social} compare prompting, sparse-autoencoder features, and
probe-based steering in economic and creative tasks. Their probe calibration
tunes one steering strength to a scalar behavioral target.
\citet{hoppe2026sliders} develop continuous personality controls that reduce
interference between traits, evaluating simultaneous shifts toward selected
high/low trait combinations. We ask a complementary question. How precisely can steering reproduce an individual’s numerical trait profile?

We calibrate the model to the measured Big Five profiles of 500 OpenPsychometrics respondents \citep{openpsychometrics2014big5}. 
The GP predicts responses at unseen doses more accurately
than the fitted linear map. In each of five passes, BO's recommendations match their intended respondents more closely, on average, than descriptive profile prompting. When each respondent can select the closest response from the full intervention--response library, BO achieves \PaperFinalBOReduction\% lower trait
mean absolute error (MAE) across the five traits than uniform random search at the same evaluation budget.

The results address three questions in turn: (i)~Can we learn the response map
(Section~\ref{sec:forward})? (ii)~How well do recommended doses fit their
intended targets (Section~\ref{sec:assigned})? (iii)~Does adaptive search
build a more useful library than additional random sampling
(Section~\ref{sec:shared})? Together they evaluate the fitting stage of
social-agent construction. Whether the resulting calibration transfers to
downstream behavior is a separate empirical question that this study leaves
open.

\section{Methods and Experimental Design}
\label{sec:methods}

\subsection{Human targets and five steering controls}
We randomly select a fixed cohort of 500 respondents from the OpenPsychometrics BIG5
dataset \citep{openpsychometrics2014big5}. Of its 19,719 records, 19,718 have
complete responses to all 50 questionnaire items. Using observed profiles
retains the combinations and correlations of traits present in these respondents,
whereas synthetic profiles generated by independently sampling five scores
would not. Respondent $p$'s five observed scores form the target vector $\yy_p\in[1,5]^5$.
All 500 profiles are known to the optimization procedure.

The instrument is the 50-item IPIP Big-Five Factor Markers questionnaire
\citep{ipip50questionnaire}, using the OpenPsychometrics wording and response
scale. The International Personality Item Pool (IPIP) is the public-domain
personality measurement resource described by \citet{goldberg2006ipip}.
Each OCEAN trait is assessed by ten questions, whose responses are scored
and averaged. We report IPIP scores in OCEAN order to match our steering controls. Using common Big Five terminology \citep{soto2017bfi2}, we label Intellect/Imagination as Openness and reverse the Emotional Stability scale so that higher scores indicate greater Neuroticism. Appendix~\ref{app:instrument} supplies the items, response
scale, and scoring key.

We use Llama 3.1 8B Instruct with fixed weights, 4-bit quantization, and
greedy questionnaire responses. Following the recursive-feature-machine (RFM)
approach of \citet{beaglehole2026steering}, we extract high- and low-pole
controllers for each of the five traits. We then keep these controllers
fixed. Their signed strengths form $\bb=(\beta_O,\beta_C,\beta_E,\beta_A,\beta_N)$:
a positive coordinate selects the high controller, and a negative coordinate
selects the low controller with the corresponding absolute strength. 

Holding the controllers fixed reduces the search over high-dimensional
activation space to five steering strengths. A single control may change several measured
traits, so we use questionnaire responses to learn how
the five strengths jointly affect the measured scores.\footnote{For an informal discussion of the difficulty of searching
high-dimensional activations, see
\href{https://huggingface.co/blog/royswastik/activation-steering}{Roy's overview
of activation steering}.} Appendix~\ref{app:controllers} details extraction and the exact
activation update.

All three steering methods use the same feasible set,
\begin{equation}
\B=[-0.3,0.3]^5.
\label{eq:box}
\end{equation}
We fixed this operating range before the present runs, motivated by earlier
exploratory observations that stronger joint interventions could destabilize
questionnaire generation. 

\subsection{Learning an intervention-to-response map}
At a dose $\bb$, the model answers each of the 50 items in a fresh
conversation. Scoring produces $\yy(\bb)$, giving the shared map
\begin{equation}
\bb\longmapsto\yy(\bb),\qquad \B\subset\R^5\longrightarrow[1,5]^5.
\label{eq:map}
\end{equation}
Steering uses exactly the same generic questionnaire prompt as the blank
baseline. The person's target does not enter that prompt, so it changes the
preferred dose rather than the map being learned.

Let $\mathcal T=\{O,C,E,A,N\}$ index the five traits. For any observed or predicted trait vector $z\in\R^5$, define
the target loss and desired intervention by
\begin{equation}
\ell_p(z)=\frac15\sum_{j\in\mathcal T}|z_j-y_{p,j}|,
\qquad
\bb_p^\star\in\arg\min_{\bb\in\B}\ell_p\bigl(\yy(\bb)\bigr).
\label{eq:loss}
\end{equation}
This objective measures similarity in the five scored characteristics. It
does not require identical answers to every item: multiple answer patterns
can yield the same trait vector.\footnote{For example, two respondents can
interchange answers of 3 and 4 on two equally keyed items and retain the
same trait mean. Equal scored traits therefore do not require identical
item responses.} Across a cohort, we report the mean of
these person-level MAEs.

To initialize learning, we draw 1,000 doses independently and
uniformly from $\B$ and freeze an 800/200 learning/test split before generation.
One learning condition is invalid, leaving $\PaperInitialFitN$ usable
intervention--response pairs in the initial library $\D_0$. All 200 test
conditions are valid, and we use them only to evaluate predictions at held-out doses.
Each library pair records a dose $\bb$ and the five trait scores $\yy(\bb)$ computed from
the model's 50 questionnaire answers.

The library serves two purposes. It provides observations to learn the
map, and it contains interventions whose responses are already known. A
simple policy is therefore to select each target's closest library response
without any further search. This gives our initial-library benchmark.

\subsection{Linear predict-then-optimize (PtO)}
The linear baseline approximates changes from the blank response
$\yy_0=\yy(0)$ by $J\bb$.\footnote{The blank condition uses the generic
respondent prompt without a personal profile or activation intervention.
Its observed score vector is $\yy_0=(3.3,3.1,3.2,3.8,3.0)$ in OCEAN
order; Appendix~\ref{app:prompts} gives the prompt.}
We fit $\widehat J$ by ordinary least squares on $\D_0$, keeping the blank
intercept fixed. The resulting prediction is
$\widehat\yy_{\rm lin}(\bb)=\yy_0+\widehat J\bb$. For each person we solve
\begin{equation}
\begin{aligned}
\min_{\bb\in\R^5}\quad &\frac15\bigl\|(\yy_p-\yy_0)-\widehat J\bb\bigr\|_1\\
\text{subject to}\quad &-0.3\leq\beta_j\leq0.3,\qquad j\in\mathcal T.
\end{aligned}
\label{eq:pto}
\end{equation}
The $\ell_1$ objective matches the reported trait MAE. The problem can be
written as a linear program and does not require inverting $\widehat J$.
We execute its minimizing dose for each person. Appendix~\ref{app:bounds}
changes only the box constraints to assess sensitivity. This provides a
simple baseline for joint calibration, although the true response map
need not be linear.

\subsection{Bayesian optimization with a shared response model}
Each evaluation requires 50 model calls, while the control space has only
five dimensions. This makes surrogate-based Bayesian optimization a suitable
approach to an expensive black-box objective \citep{frazier2018tutorial}.
Recent work also uses BO to allocate evaluations in black-box LLM prompt
selection \citep{schneider2025hbbops}. Here it uses observed trait responses
and uncertainty to identify interventions that could improve an existing match.

We place a Gaussian-process prior on each component of the unknown response
map, retaining the notation $y_j$:
\begin{equation}
y_j:\B\longrightarrow\R,\qquad
y_j\sim\operatorname{GP}(m_j,k_j),\quad j\in\mathcal T.
\label{eq:gp}
\end{equation}
Each GP uses \emph{all five} intervention coordinates. Evaluating them at
the same dose gives one joint prediction,
$\yy(\bb)=(y_O(\bb),y_C(\bb),y_E(\bb),y_A(\bb),y_N(\bb))^\top$.
We fit the five output processes independently. Learning estimates the
shared map, optimization then chooses a single $\bb$ for each person's
joint target.
Constant means and Mat\'ern kernels give a locally smooth statistical
approximation \citep{rasmussen2006gpml}. 

Let $\D_{r-1}$ contain the initial library and valid BO responses observed
before pass $r$. Conditioning gives a posterior over response functions,
\begin{equation}
y_j\mid\D_{r-1}\sim\operatorname{GP}(m_{j,r},k_{j,r}),\qquad j\in\mathcal T.
\label{eq:posterior}
\end{equation}
At an unevaluated dose this supplies a predicted response and uncertainty.
The posterior mean can be used to predict scores; choosing the next
experiment additionally involves the possibility of improving the best
response already observed.

For each person, let $b_{p,r}$ be the smallest observed loss in
$\D_{r-1}$. We recommend a dose by expected improvement (EI)
\citep{jones1998efficient}:
\begin{equation}
\alpha_{p,r}(\bb)
=\E\!\left[\left(b_{p,r}-\ell_p(\yy(\bb))\right)_+
\mid\D_{r-1}\right],
\qquad
\widehat\bb_{p,r}\approx\arg\max_{\bb\in\B}\alpha_{p,r}(\bb),
\label{eq:ei}
\end{equation}
where $(u)_+=\max(u,0)$. Possible responses that improve the current match
contribute their gain; worse responses contribute zero because the previous
match remains available. Alternatively, one might be tempted to optimize using posterior mean directly. However, choosing the dose whose posterior mean response is closest to the target overlooks uncertainty in that prediction. EI instead accounts for both the chance
of improving the current match and the size of the possible gain.\footnote{A
mean response can look close to the target because deviations on opposite
sides cancel. For example, equally likely responses of 3 and 5 have mean 4,
but both miss a target of 4 by one point. Averaging responses before computing
loss therefore differs from averaging losses. EI goes a step further and
averages only improvements over the best observed match.}

\begin{figure}[htbp]
\centering
\begin{tikzcd}[ampersand replacement=\&,column sep=large,
every label/.append style={font=\small}]
\D_{r-1}\arrow[r,"\text{fit}"]\&
\yy\mid\D_{r-1}\arrow[r,"\text{EI}"]\&
\{\widehat\bb_{p,r}\}_{p=1}^{500}\arrow[r,"\text{execute}"]\&\D_r
\end{tikzcd}
\caption{Five passes of shared learning.}
\label{fig:passes}
\end{figure}

We repeat this procedure for five passes, with one new recommendation per
person per pass (Figure~\ref{fig:passes}). Each pass uses a fixed posterior based on the observations available at its start, so that everyone’s recommendation uses the same information.
Only after the pass do we pool the new responses and refit, including
re-estimation of kernel parameters. A final refit after pass 5 is used only
to assess forward prediction. Appendix~\ref{app:computation} summarizes
computation and evaluation costs.

\subsection{Comparators and evaluation rules}
We compare three ways to choose steering doses: BO, linear PtO, and uniform random search. Uniform random search provides a budget-matched baseline for adaptive
optimization \citep{bergstra2012random}. It draws independently from $\B$,
assigning one dose to each of 500 people in each of five passes. BO receives
the same budget: $5\times500=2{,}500$ additional interventions, each requiring
50 item calls, or 125,000 calls per method. Both may use the same initial
library for final selection; their new responses remain separate. As a
one-shot method, PtO contributes a single 500-dose batch.

We also evaluate two prompting baselines. Blank prompting asks the model to answer as a generic respondent and
produces the baseline trait vector $\yy_0$. Profile prompting
instead supplies a paragraph describing the target's behavioral tendencies.
The paragraph is generated from fixed score-to-behavior bands, with constraints
against adding unsupported personal details. Its numerical target is not
printed in the questionnaire prompt. These are direct prompting benchmarks,
not prompt-search procedures matched to BO's budget.
Appendix~\ref{app:prompts} gives the exact templates and an example.

We distinguish two calibration evaluations. \emph{Assigned-dose evaluation}
scores only the new intervention recommended for the intended person:
\begin{equation}
\operatorname{MAE}_{\rm assigned}(r)
=\frac1{|\mathcal V_r|}\sum_{p\in\mathcal V_r}
\ell_p\bigl(\yy(\widehat\bb_{p,r})\bigr),
\label{eq:assigned}
\end{equation}
where $\mathcal V_r$ contains people with a valid assigned response.
\emph{Shared-library evaluation} lets each of the $n=500$ people select the closest valid
response from the collection $\D$ available up to that point, including
earlier evaluations and other people's doses:
\begin{equation}
\operatorname{MAE}_{\rm lookup}(\D)
=\frac1n\sum_{p=1}^n\min_{(\bb,y)\in\D}\ell_p(y).
\label{eq:lookup}
\end{equation}
For the latter, every method starts with $\D_0$ and adds only its own
responses. The first evaluation measures recommendation quality without
shared final selection; BO still shares data for learning in both.

Malformed item answers are not retried under greedy decoding. A condition
is valid only when all 50 items can be scored; invalid attempts consume the
budget and are excluded from learning and selection. We report each
assigned-dose average together with the number of valid responses. When
coverage differs, we also compare methods on the same people with valid
responses under both methods.
The 200-dose prediction test compares predicted scores with measured LLM
scores at already specified doses, rather than optimizing doses for new
human targets. Further execution and budget details are in
Appendix~\ref{app:computation}.

\section{Results}
\label{sec:results}

\subsection{Can we predict responses to unseen interventions?}
\label{sec:forward}
The initial 1,000-intervention experiment reserved 200 doses for this test,
separate from the $\PaperInitialFitN$ valid observations in $\D_0$. At each
held-out dose $\bb_i$, we predict the LLM's trait response using
\begin{equation}
\widehat\yy_{\rm lin}(\bb_i)=\yy_0+\widehat J\bb_i,
\qquad
\widehat\yy_{\rm GP}(\bb_i)=\E[\yy(\bb_i)\mid\D].
\label{eq:forward-predictions}
\end{equation}
Here $\D$ is the fitting library for the GP being evaluated. The input dose
is already given, so prediction requires neither EI search nor new
questionnaire responses.

\begin{figure}[!htbp]
\centering
\includegraphics[width=\linewidth]{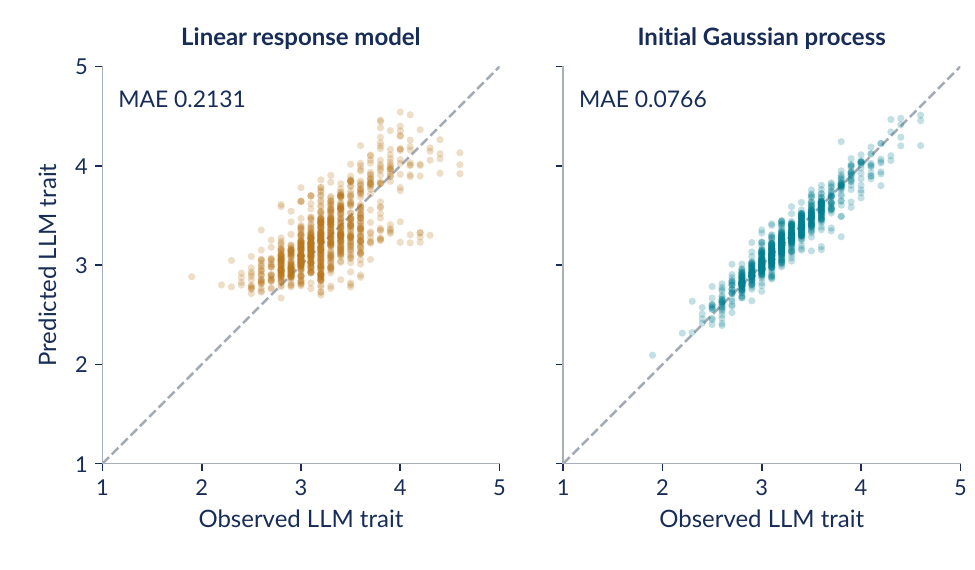}
\caption{Predicted versus observed LLM traits at 200 held-out doses.
Both models use the same initial fitting library. The diagonal marks
perfect prediction; each panel contains 200 doses $\times$ five traits.}
\label{fig:forward}
\end{figure}
\FloatBarrier

The initial GP tracks the observed responses more closely than the linear
map (Figure~\ref{fig:forward}). Its MAE is $\PaperInitialGPMAE$, compared
with $\PaperLinearForwardMAE$ for the linear approximation. Refitting the
GP on $\PaperFinalFitN$ observations after all five BO passes lowers MAE
to $\PaperFinalGPMAE$ (Table~\ref{tab:forward}). The largest prediction
gain therefore comes from the initial nonlinear model, with a smaller
further improvement from the additional observations.

\begin{table}[htbp]
\centering\small
\PaperForwardTable
\caption{Prediction error on the held-out doses, averaged across five
traits. The no-change benchmark predicts $\yy_0$ at every dose. The final
GP is refitted after pass 5 and is not used to recommend pass-5 doses.}
\label{tab:forward}
\end{table}

The GP predicts traits at held-out doses more accurately than the linear baseline. Next, we evaluate how closely BO’s recommended doses match their intended targets by administering the questionnaire at each proposed dose.

\subsection{Do recommended doses match their intended human targets?}
\label{sec:assigned}
We score only each person's newly assigned response~\eqref{eq:assigned}. Table~\ref{tab:assigned} reports every BO and random
pass, together with the single-run baselines.

\begin{table}[htbp]
\centering\small\setlength{\tabcolsep}{5pt}
\PaperAssignedTable
\caption{Mean trait MAE for newly assigned responses; lower is better.
All BO and random passes have 500 valid conditions. PtO has 459 valid
conditions; its MAE is conditional on those people. Prompting provides
direct-response references, not equally budgeted search procedures.}
\label{tab:assigned}
\end{table}

BO's fresh recommendations outperform profile prompting in every pass.
Their mean MAEs remain between 0.4128 and 0.4261, while uniform draws range
from 0.7271 to 0.7319. The fifth-pass value, $\PaperBOFinalMAE$, is
$\PaperBOFinalProfileReduction\%$ below the profile-prompt value of
$\PaperProfileMAE$. Appendix~\ref{app:bounds} shows that smaller PtO bounds improve its results but do not reverse the assigned-dose comparison.

These averages do not imply improvement for every person. Relative to
profile prompting, the pass-5 BO recommendation improves
$\PaperAssignedImprovedN$ of 500 targets, ties for
$\PaperAssignedTiedN$, and worsens for $\PaperAssignedWorsenedN$.
The left panel of Figure~\ref{fig:individual} plots each person's
profile-prompt MAE minus their assigned BO MAE. Positive differences mean
that BO fits that person better; larger values indicate greater improvement.

Despite BO's adaptive updates, fresh recommendation quality does not
improve monotonically across passes. EI seeks a chance of beating an incumbent, and a newly evaluated dose need
not do so. The stable assigned-dose averages therefore support the quality
of BO recommendations relative to these baselines, while the next comparison
asks what the additional evaluations contribute to the available library.

\subsection{Does BO outperform collecting more random responses?}
\label{sec:shared}
Shared lookup~\eqref{eq:lookup} is a strong baseline. Even before further search, choosing
each target's closest response among the initial $\PaperInitialFitN$
conditions gives MAE $\PaperInitialLookupMAE$, below descriptive profile
prompting and the average fresh BO recommendation. To isolate the value of
search, we now give every steering method the same initial library and the
same shared-selection rule.

\begin{figure}[!htbp]
\centering
\includegraphics[width=\linewidth]{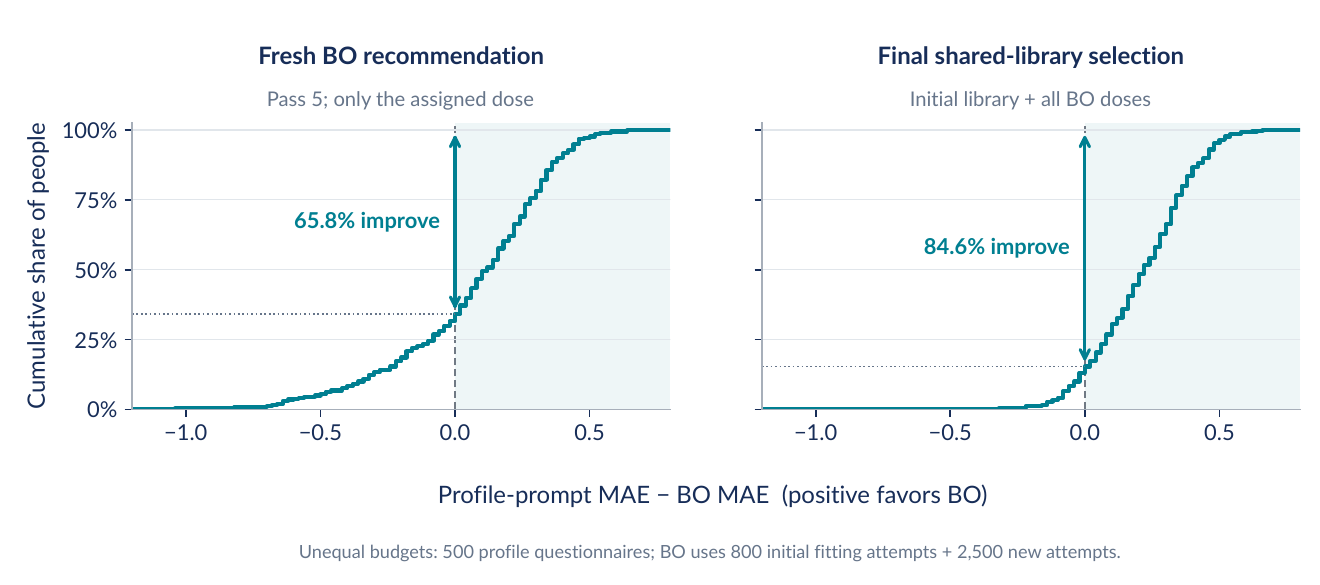}
\caption{Empirical cumulative distributions of person-level gains over
profile prompting. The horizontal axis is profile-prompt MAE minus BO MAE;
positive values favor BO. Left: the assigned pass-5 dose
(Section~\ref{sec:assigned}). Right: the closest response in the final BO
library (Section~\ref{sec:shared}). Shading marks positive gains; vertical
brackets show the fraction improved, $1-F(0)$, where $F$ is the empirical CDF.}
\label{fig:individual}
\end{figure}

Figure~\ref{fig:individual} contrasts assigned doses with shared selection
at the end of BO. With the final library available, $\PaperSharedImprovedN$
targets improve over their own profile-prompt responses,
$\PaperSharedTiedN$ tie, and $\PaperSharedWorsenedN$ worsen (right panel).

\begin{table}[htbp]
\centering\small
\PaperSharedTable
\caption{Shared-library trait MAE on all 500 people. Each column includes
the same initial fitting library and only that method's additional valid
responses. The zero-addition row is identical by construction. Budgets count attempted interventions, including invalid ones.
PtO adds one batch. For context, assigned blank and profile prompting have
MAEs $\PaperBlankMAE$ and $\PaperProfileMAE$, respectively.}
\label{tab:shared}
\end{table}

\begin{figure}[htbp]
\centering
\includegraphics[width=\linewidth]{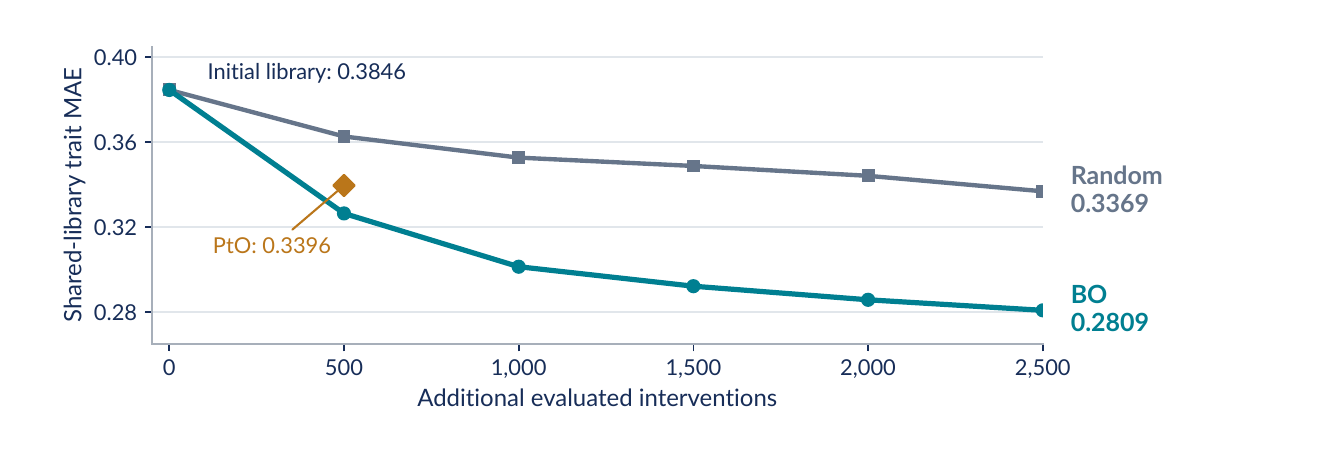}
\caption{Calibration gained as the shared library grows. BO and random
search have identical starting observations, bounds, and selection rules.
PtO is shown at its single 500-dose budget. The initial collection required
1,000 evaluations, of which 200 are reserved for prediction testing and
never enter these libraries.}
\label{fig:budget}
\end{figure}

Under shared lookup, BO yields lower mean trait MAE than uniform random
search at every budget tested, from 500 to 2,500 additional interventions
(Figure~\ref{fig:budget}). With just 500 interventions beyond the initial
library, BO achieves an MAE of $\PaperBOFirstLookupMAE$, below both PtO
at the same budget ($\PaperPtOLookupMAE$) and random search even at 2,500
interventions ($\PaperRandomFinalLookupMAE$). At the full budget, BO reaches
$\PaperBOFinalLookupMAE$, a $\PaperFinalBOReduction\%$ reduction relative
to random search (Table~\ref{tab:shared}).

\paragraph{Robustness.}
BO's assigned-dose advantage holds across ten fresh libraries of 100 doses,
each evaluated with 50 fixed targets and one recommendation pass, and
against smaller-box PtO (Appendices~\ref{app:bounds}
and~\ref{app:initialization}). BO also beats random search under shared
lookup in all ten repetitions, although PtO yields the better shared
library in nine; this reversal occurs in the smaller-budget setting.

\section{Discussion}
\label{sec:discussion}
The experiment identifies a useful role for statistical calibration in
constructing social agents. Extracted activation directions provide controls,
but their strengths need to be fitted to the desired joint response. A
shared nonlinear surrogate predicts this response substantially better
than the tested linear approximation. BO then produces useful individual
recommendations and, in the main experiment, builds a more effective
response library than equally many uniform evaluations.

Reuse is central to this design. Once a response has been observed, every
target can be compared with it at negligible additional generation cost.
This makes calibration across a cohort different from running unrelated
optimization problems for each person. It also makes the selection rule
part of the method: an intervention that poorly matches its intended target
can still improve the library for somebody else. The small-library
experiment demonstrates that these two forms of usefulness can lead to
different method rankings.

The principal boundary is between calibration and behavioral transfer.
Trait scores aggregate several answers, so people with the same scores can
differ on individual items. Matching those scores is a coherent fitting
objective, but does not establish that an agent reproduces preferences,
choices, or interactions beyond the questionnaire. Testing that next step
requires freezing the calibrated intervention and evaluating it on an
independently chosen instrument or economic task.

We demonstrate joint calibration with one language model and one personality questionnaire, using descriptive profile prompting as the benchmark. The same question arises when a simulation calls for agents with specified attitudes toward risk or judgments of fairness. Extending the method to those settings would require human measurements and suitable steering controls. Our contribution is a way to check, before a simulation begins, how closely each agent matches the profile it was assigned.

\FloatBarrier
\begin{spacing}{1.05}
\bibliographystyle{plainnat}
\bibliography{references}
\end{spacing}
\clearpage
\appendix
\begin{spacing}{1.1}
\section{Sensitivity to the PtO Feasible Region}
\label{app:bounds}
For each $b\in\{0.3,0.2,0.1\}$, we re-optimize the same fitted linear
response model:
\begin{equation}
\begin{aligned}
\underset{\bb\in\R^5}{\operatorname{minimize}}\quad
&\frac15\left\|(\yy_p-\yy_0)-\widehat J\bb\right\|_1\\
\text{subject to}\quad
&-b\leq\beta_j\leq b,\qquad j\in\mathcal T.
\end{aligned}
\label{eq:pto-bounds}
\end{equation}
Only the constraint changes. The fitted $\widehat J$, blank intercept,
500 targets, controllers, and generation settings stay fixed. Table~\ref{tab:bounds} summarizes the results.

\begin{table}[htbp]
\centering\small
\PaperBoundsTable
\caption{PtO bound sensitivity. At each bound, we compute assigned-dose MAE
(Equation~\ref{eq:assigned}) using only the observed response to each
respondent's recommended dose. The matched column uses the same 459 people
with valid responses under every bound. Blank and profile prompting provide
references on both the full and matched cohorts.}
\label{tab:bounds}
\end{table}

Smaller boxes improve realized MAE and remove the invalid responses observed
at $\pm0.3$. Even the smallest tested box remains above profile prompting.
Therefore, the main bound should be understood as a common operating range,
not an optimal or uniformly stable choice.

\section{Sensitivity to the Initial Library}
\label{app:initialization}
We draw ten fresh uniform libraries of 100 interventions each and use the
same fixed 50-person subset throughout. Within each repetition, PtO and BO
learn from that repetition's library; all three methods then execute one
new recommendation per person. BO has one fit and recommendation pass.
Every initial library
and every BO and random recommendation is valid; PtO coverage is reported
in Table~\ref{tab:initialization}.

\begin{table}[htbp]
\centering\footnotesize
\setlength{\tabcolsep}{5pt}
\PaperRobustnessTable
\caption{Ten initial-library repetitions. Each cell reports mean trait MAE.
Parentheses give valid PtO recommendations out of 50; all other assigned
cells have 50 valid responses. Shared lookup evaluates all 50 targets using
the repetition's initial library plus that method's valid new responses.
The last row averages the ten replicate means.}
\label{tab:initialization}
\end{table}

\begin{figure}[htbp]
\centering
\includegraphics[width=.96\linewidth]{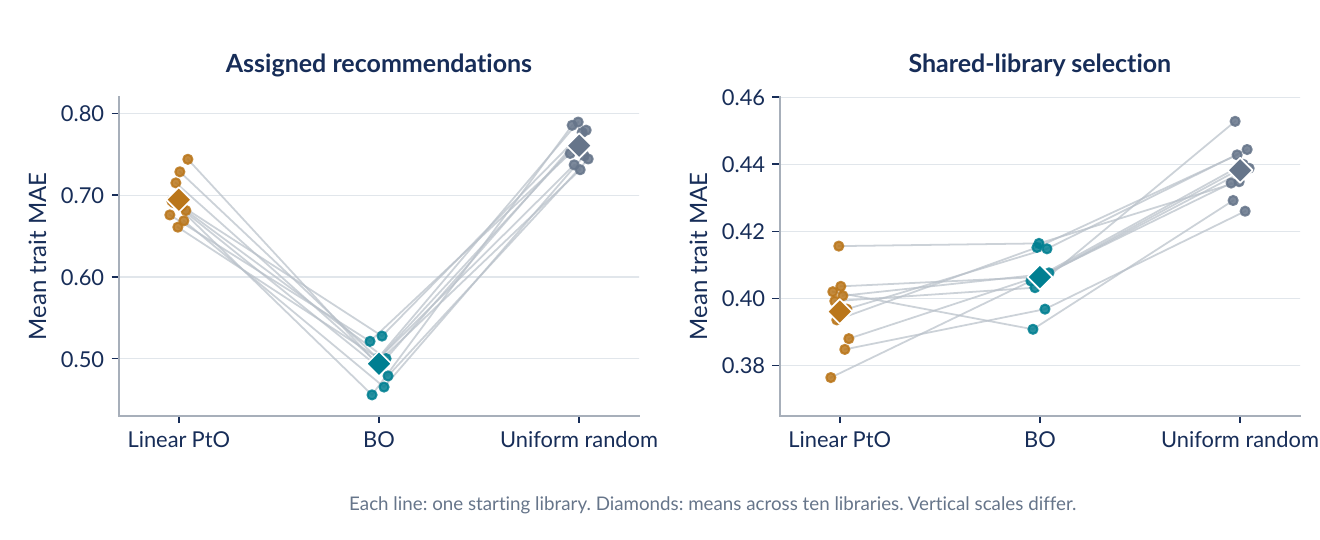}
\caption{Recommendation quality and shared-library value across ten fresh
initializations. Connecting lines identify the same repetition across
methods. The two selection rules answer different questions and can rank
methods differently.}
\label{fig:initialization}
\end{figure}

BO has lower assigned-dose MAE \eqref{eq:assigned} than PtO and random search in all ten
repetitions; the PtO comparison also holds on matched valid targets. Under
shared lookup \eqref{eq:lookup}, BO beats random search in all ten, but PtO beats BO in nine.
Figure~\ref{fig:initialization} makes this change in ranking explicit.
An intervention that is a poor match for its intended target can still be
useful to another person in the cohort.

This exercise studies initialization under a smaller budget. Relative to
the main experiment, it changes the library size, cohort size, number of
additional evaluations, and update horizon. It therefore does not isolate
cohort size as the cause of the shared-lookup reversal. Whether larger
repetitions recover the main ranking remains an empirical question.

\FloatBarrier
\section{Controller Construction and Activation Intervention}
\label{app:controllers}
We use the RFM extraction pipeline of
\citet{beaglehole2026steering}. The 15 BFI-2 facets provide a scaffold for
OCEAN concept statements \citep{soto2017bfi2}.
For each of five domains, three facets, and two poles, the local model
generates 40 first-person statements: 1,200 total. Mechanical checks reject
malformed, repetitive, or unsuitable wording, and a same-model semantic
judge checks the intended facet and pole. The statements describe generic
behavioral tendencies; generation uses neither the human targets nor their
IPIP answers.\footnote{For Openness, Intellectual Curiosity, high pole, one
accepted statement is: ``I tend to seek out abstract ideas and concepts
that expand my understanding of the world.''} Appendix~\ref{app:concept-prompts}
provides the facet scaffold and generation template.

Each trait has a fixed facet-stratified split: 192 training statements and
48 validation statements, comprising 32 and eight per facet/pole,
respectively. High and low controllers use the same split with reversed
binary labels. RFM fitting uses up to eight iterations and validation AUC for
selection. Five leading components of the average gradient outer product
are extracted and oriented toward the labeled pole; execution uses the
first component. These discrimination diagnostics do not establish that
each controller affects only its named trait.

Let $v_{\ell,j}^{\rm high}$ and $v_{\ell,j}^{\rm low}$ denote the first
components for trait $j$ at layer $\ell$. With $(a)_+=\max(a,0)$,
the implemented update is
\begin{equation}
h_\ell\longmapsto h_\ell+\Delta h_\ell(\bb),\qquad
\Delta h_\ell(\bb)=\sum_{j\in\mathcal T}
\left[(\beta_j)_+v_{\ell,j}^{\rm high}
+(-\beta_j)_+v_{\ell,j}^{\rm low}\right].
\label{eq:pole-update}
\end{equation}
The same five strengths apply at the last 30 of Llama's 32 transformer
blocks; the first two are unmodified. The addition is broadcast across
token positions during inference. We do not
renormalize the combined vector. The low-pole vector is separately learned
and need not be the negative of the high-pole vector. Thus five signed
parameters specify the intervention even though ten pole-specific
controllers are stored.

\section{Computation}
\label{app:computation}
The experiments run locally on an NVIDIA GeForce RTX 4070 Ti. We use
Llama 3.1 8B Instruct with 4-bit NF4 double quantization
and bfloat16 computation. Questionnaire calls are greedy, allow at most
32 generated tokens, and start a fresh conversation per item. All 50 items
are attempted; malformed responses are recorded without identical retries.
Complete conditions are required for trait scoring.

GP regression uses standard Gaussian conditioning with constant means,
Mat\'ern-$3/2$ kernels, and hyperparameters estimated by marginal likelihood
\citep{rasmussen2006gpml}. EI is evaluated from posterior draws and maximized
numerically over candidate interventions \citep{frazier2018tutorial}.
These calculations run on the CPU; questionnaire generation accounts for
the main computational cost. Table~\ref{tab:budgets} summarizes the computational costs.

\begin{table}[htbp]
\centering\small
\setlength{\tabcolsep}{4.5pt}
\begin{tabular}{@{}lrrr@{}}
\toprule
Stage & Questionnaires & Item calls & Elapsed hours\\
\midrule
Blank & 1 & 50 & $<0.01$\\
Profile prompting & 500 & 25,000 & 2.19\\
Initial library (including forward tests) & 1,000 & 50,000 & 4.30\\
Linear PtO & 500 & 25,000 & 2.14\\
BO, five passes & 2,500 & 125,000 & 10.20\\
Uniform random, five passes & 2,500 & 125,000 & 10.50\\
\midrule
Main experiment & 7,001 & 350,050 & 29.32\\
Ten-library sensitivity & 2,500 & 125,000 & 10.30\\
Two additional PtO bounds & 1,000 & 50,000 & 4.16\\
\bottomrule
\end{tabular}
\caption{Questionnaire budgets and observed elapsed time. Counts include
invalid attempts. Times run from the first to last recorded item answer
within each run and are summed across runs for aggregate rows; intervening
BO updates are included. Model loading, initial fitting, and concept/profile
generation are excluded. The initial library allocates 40,000 calls to
learning and 10,000 to forward testing.}
\label{tab:budgets}
\end{table}

\FloatBarrier
\section{IPIP-50 Items and Scoring}
\label{app:instrument}
Table~\ref{tab:items} reproduces the questionnaire used throughout, in its
presentation order. Responses use a five point scale with
$1=$ Disagree, $3=$ Neutral, and $5=$ Agree.
For response $a$, a positive key contributes $a$ and a negative key
contributes $6-a$. Each trait is the mean of its ten contributions.
The Neuroticism key already incorporates the reversal from Emotional
Stability. Items and scoring follow the official IPIP 50-item questionnaire
\citep{ipip50questionnaire}, with the first-person wording and
Disagree--Agree response anchors used by OpenPsychometrics.

{\small\setlength{\tabcolsep}{5pt}
\begin{longtable}{@{}r p{.73\linewidth} cc@{}}
\caption{Canonical questionnaire items and OCEAN scoring keys.}
\label{tab:items}\\
\toprule
No. & Item & Trait & Key \\
\midrule
\endfirsthead
\multicolumn{4}{l}{\textit{Table \thetable\ continued}}\\
\toprule
No. & Item & Trait & Key \\
\midrule
\endhead
\midrule
\multicolumn{4}{r}{\textit{Continued on next page}}\\
\endfoot
\bottomrule
\endlastfoot
1 & I am the life of the party. & E & $+$ \\
2 & I feel little concern for others. & A & $-$ \\
3 & I am always prepared. & C & $+$ \\
4 & I get stressed out easily. & N & $+$ \\
5 & I have a rich vocabulary. & O & $+$ \\
6 & I don't talk a lot. & E & $-$ \\
7 & I am interested in people. & A & $+$ \\
8 & I leave my belongings around. & C & $-$ \\
9 & I am relaxed most of the time. & N & $-$ \\
10 & I have difficulty understanding abstract ideas. & O & $-$ \\
11 & I feel comfortable around people. & E & $+$ \\
12 & I insult people. & A & $-$ \\
13 & I pay attention to details. & C & $+$ \\
14 & I worry about things. & N & $+$ \\
15 & I have a vivid imagination. & O & $+$ \\
16 & I keep in the background. & E & $-$ \\
17 & I sympathize with others' feelings. & A & $+$ \\
18 & I make a mess of things. & C & $-$ \\
19 & I seldom feel blue. & N & $-$ \\
20 & I am not interested in abstract ideas. & O & $-$ \\
21 & I start conversations. & E & $+$ \\
22 & I am not interested in other people's problems. & A & $-$ \\
23 & I get chores done right away. & C & $+$ \\
24 & I am easily disturbed. & N & $+$ \\
25 & I have excellent ideas. & O & $+$ \\
26 & I have little to say. & E & $-$ \\
27 & I have a soft heart. & A & $+$ \\
28 & I often forget to put things back in their proper place. & C & $-$ \\
29 & I get upset easily. & N & $+$ \\
30 & I do not have a good imagination. & O & $-$ \\
31 & I talk to a lot of different people at parties. & E & $+$ \\
32 & I am not really interested in others. & A & $-$ \\
33 & I like order. & C & $+$ \\
34 & I change my mood a lot. & N & $+$ \\
35 & I am quick to understand things. & O & $+$ \\
36 & I don't like to draw attention to myself. & E & $-$ \\
37 & I take time out for others. & A & $+$ \\
38 & I shirk my duties. & C & $-$ \\
39 & I have frequent mood swings. & N & $+$ \\
40 & I use difficult words. & O & $+$ \\
41 & I don't mind being the center of attention. & E & $+$ \\
42 & I feel others' emotions. & A & $+$ \\
43 & I follow a schedule. & C & $+$ \\
44 & I get irritated easily. & N & $+$ \\
45 & I spend time reflecting on things. & O & $+$ \\
46 & I am quiet around strangers. & E & $-$ \\
47 & I make people feel at ease. & A & $+$ \\
48 & I am exacting in my work. & C & $+$ \\
49 & I often feel blue. & N & $+$ \\
50 & I am full of ideas. & O & $+$ \\
\end{longtable}}

\section{Questionnaire Prompts and Profile Construction}
\label{app:prompts}
The following templates are used for every questionnaire item.
\texttt{\{item\_text\}} is replaced by its statement in
Table~\ref{tab:items}; \texttt{\{profile\}} is replaced by the saved
behavioral paragraph. Steering uses the blank template.

\paragraph{System message, all methods.}
\begin{Prompt}
You answer questionnaire items carefully. Return only a JSON object with one integer field named answer.
\end{Prompt}

\paragraph{Blank and steered user message.}
\begin{Prompt}
Complete this personality questionnaire item as a generic respondent.
Do not assume any particular personal profile. Use the full response scale when appropriate, and do not default to agreement or an extreme.

Use this response scale:
1 = Disagree
2 = Between Disagree and Neutral
3 = Neutral
4 = Between Neutral and Agree
5 = Agree

Statement:
{item_text}

Return JSON only, in this exact schema:
{"answer": <one integer from 1 to 5>}
\end{Prompt}

\Needspace{0.45\textheight}
\paragraph{Profile-prompted user message.}
\begin{Prompt}
You have the following behavioral profile:

{profile}

Complete this personality questionnaire item as yourself, using the profile as the basis for your self-description. Use the full response scale when appropriate, and do not default to agreement or an extreme.

Use this response scale:
1 = Disagree
2 = Between Disagree and Neutral
3 = Neutral
4 = Between Neutral and Agree
5 = Agree

Statement:
{item_text}

Return JSON only, in this exact schema:
{"answer": <one integer from 1 to 5>}
\end{Prompt}

\paragraph{From numerical scores to a behavioral paragraph.}
Each trait score maps to one of six fixed descriptions: low ($<2$),
somewhat low ($[2,2.75)$), mixed ($[2.75,3.25)$), somewhat elevated
($[3.25,4)$), elevated ($[4,4.5)$), or very elevated ($\geq4.5$).
The local model joins the resulting five behavioral facts into five
sentences, in OCEAN order, within a 90--130-word paragraph. Instructions
prohibit invented biography, numerical scores, trait labels, and changes
to the stated tendencies. Mechanical checks and a semantic judge assess
fidelity; unsuccessful drafts receive revised correction prompts, with a
fixed-template fallback after three attempts. These profile-generation
checks are separate from the no-retry questionnaire protocol.

For the target $(4.5,4.9,3.5,3.8,4.6)$, the saved 106-word profile is:
\begin{quote}\small
This person tends to approach problems and situations with an intellectual curiosity that drives them to explore novel and abstract perspectives, often going beyond conventional approaches. They are structured and dependable in their planning and follow-through, consistently attending to details and sustaining effort until tasks are complete. In social situations, this person is generally engaged and comfortable initiating interaction, but also values quieter or more independent settings. They tend to be considerate and cooperative in their interpersonal interactions, while also being direct when necessary. When faced with stress or uncertainty, this person is highly sensitive to pressure and may experience intense worry, discouragement, or emotional disruption.
\end{quote}
The benchmark therefore asks the model to realize a descriptive profile;
it does not supply the optimizer's exact numerical target in the prompt.

\section{Concept-Statement Generation Prompts}
\label{app:concept-prompts}
The training statements in Appendix~\ref{app:controllers} use the following
BFI-2 facet scaffold \citep{soto2017bfi2}. Each facet is generated at both high and low poles,
with 40 accepted statements per facet and pole.

\begin{center}
\small
\begin{tabular}{@{}lp{.76\linewidth}@{}}
\toprule
Domain & Facets\\
\midrule
O & Intellectual Curiosity; Aesthetic Sensitivity; Creative Imagination\\
C & Organization; Productiveness; Responsibility\\
E & Sociability; Assertiveness; Energy Level\\
A & Compassion; Respectfulness; Trust\\
N & Anxiety; Depression; Emotional Volatility\\
\bottomrule
\end{tabular}
\end{center}

The system message and user template below generate one batch of up to 16
candidates. Braced fields are replaced by the domain, facet, pole, and
corresponding behavioral definitions. For example, the high Intellectual
Curiosity tendency is ``seeking abstract ideas, explanations, and conceptual
exploration''; its contrast is ``preferring concrete, practical information
over abstract exploration.'' On later batches, the optional retry note
lists previously used sentence openings to avoid. Generation continues
until 40 statements pass the mechanical and semantic checks.

\paragraph{System message.}
\begin{Prompt}
You generate concise, diverse, and evaluatively neutral training statements from structured OCEAN definitions. Return only valid JSON.
\end{Prompt}

\Needspace{0.55\textheight}
\paragraph{User template.}
\begin{Prompt}
Generate {requested} concise first-person behavioral-principle statements.

Concept dataset: {concept}
OCEAN dimension: {trait_name}
BFI-2 domain name: {bfi2_domain_name}
Class: {pole}
Class meaning: {pole_meaning}.

Facet theme: {facet_name}
Target tendency: {target_tendency}.
Opposite tendency for contrast only: {contrast_tendency}.
{retry_note}
Requirements:
- express only the target tendency, not the opposite tendency
- 8 to 24 words each
- one sentence each, ending with a period
- first-person phrasing using I, me, or my
- high-level behavioral principles, preferences, or recurring tendencies
- no questionnaire answers, scores, scenarios, examples, dialogue, advice, or commands
- describe both poles neutrally; neither pole is morally, clinically, or socially better
- for the low class, express a coherent alternative preference rather than merely negating the high class
- do not mention OCEAN, Big Five, personality, scores, tests, dimensions, traits, poles, or questionnaires
- do not use the dimension name, domain name, or facet name directly
- avoid clinical labels, diagnoses, moral judgments, and extreme words such as always or never
- stay within the specified facet rather than adding characteristics from other OCEAN dimensions
- vary wording and sentence openings

Return only valid JSON:
{"items": [{"text": "..."}]}
\end{Prompt}
\end{spacing}

\end{document}